\documentclass[letterpaper, 10 pt, conference]{ieeeconf}  

\IEEEoverridecommandlockouts     

\usepackage{amsmath,amsfonts}
\usepackage{algorithmic}
\usepackage{algorithm}
\usepackage{array}
\usepackage[caption=false,font=normalsize,labelfont=rm,textfont=rm]{subfig}
\usepackage{overpic}
\usepackage{textcomp}
\usepackage{stfloats}
\usepackage{url}
\usepackage{verbatim}
\usepackage{graphicx}
\usepackage{cite}
\usepackage{bbm}
\usepackage{times}
\usepackage{soul,xcolor}
\usepackage{url}
\definecolor{cvprblue}{rgb}{0.21,0.49,0.74}
\definecolor{LightCyan}{rgb}{0.88,1,1}
\usepackage[colorlinks,citecolor=cvprblue,linkcolor=red]{hyperref}
\usepackage[capitalize,noabbrev]{cleveref}

\usepackage{booktabs}

\usepackage{caption}

\title{\LARGE \bf
Uncertainty-aware Probabilistic 3D Human Motion Forecasting \\ via Invertible Networks
}

\author{Yue Ma\textsuperscript{1}, 
        Kanglei Zhou\textsuperscript{1}, 
        Fuyang Yu\textsuperscript{1}, 
        Frederick W. B. Li\textsuperscript{2}, and 
        Xiaohui Liang\textsuperscript{1,3,*}
\thanks{\textsuperscript{1} The State Key Laboratory of Virtual Reality Technology and Systems, Beihang University, Beijing, China}
\thanks{\textsuperscript{2} Department of Computer Science, Durham University, Durham, UK}
\thanks{\textsuperscript{3} Zhongguancun Laboratory, Beijing, China}
\thanks{\textsuperscript{*} Corresponding author: Xiaohui Liang \textit{liang\_xiaohui@buaa.edu.cn}}
\thanks{This work was supported by the National Natural Science Foundation of China under Project 62272019.}
}

\begin{document}
\maketitle
\thispagestyle{empty}
\pagestyle{empty}

\begin{abstract}
3D human motion forecasting aims to enable autonomous applications. Estimating uncertainty for each prediction (\emph{i.e.}, confidence based on probability density or quantile) is essential for safety-critical contexts like human-robot collaboration to minimize risks. However, existing diverse motion forecasting approaches struggle with uncertainty quantification due to implicit probabilistic representations hindering uncertainty modeling. We propose ProbHMI, which introduces invertible networks to parameterize poses in a disentangled latent space, enabling probabilistic dynamics modeling. A forecasting module then explicitly predicts future latent distributions, allowing effective uncertainty quantification. Evaluated on benchmarks, ProbHMI achieves strong performance for both deterministic and diverse prediction while validating uncertainty calibration, critical for risk-aware decision making.
\end{abstract}

\section{Introduction}
\begin{figure}[htb]\captionsetup[subfloat]{font=footnotesize}
	\centering
        \subfloat[Generative model-based methods]{
            \includegraphics[height=0.4\linewidth]{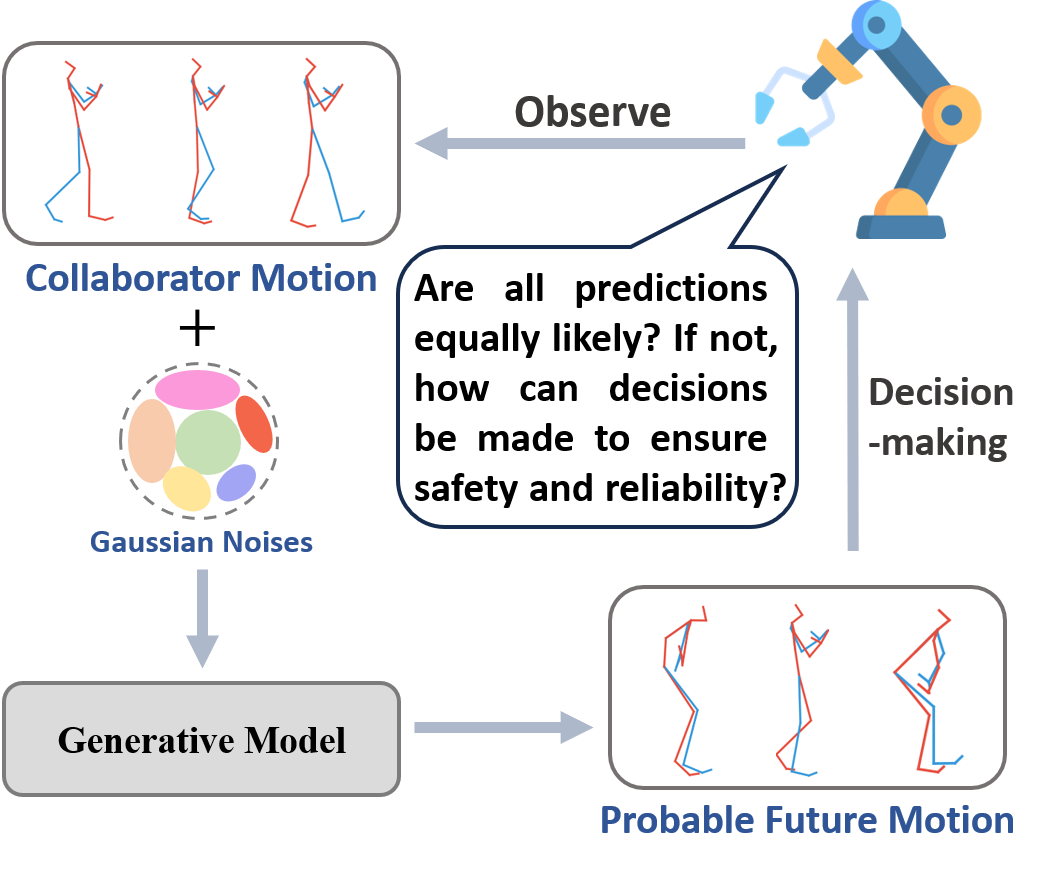}
            \label{fig:diverse}
        }
        \subfloat[Ours]{
            \includegraphics[height=0.4\linewidth]{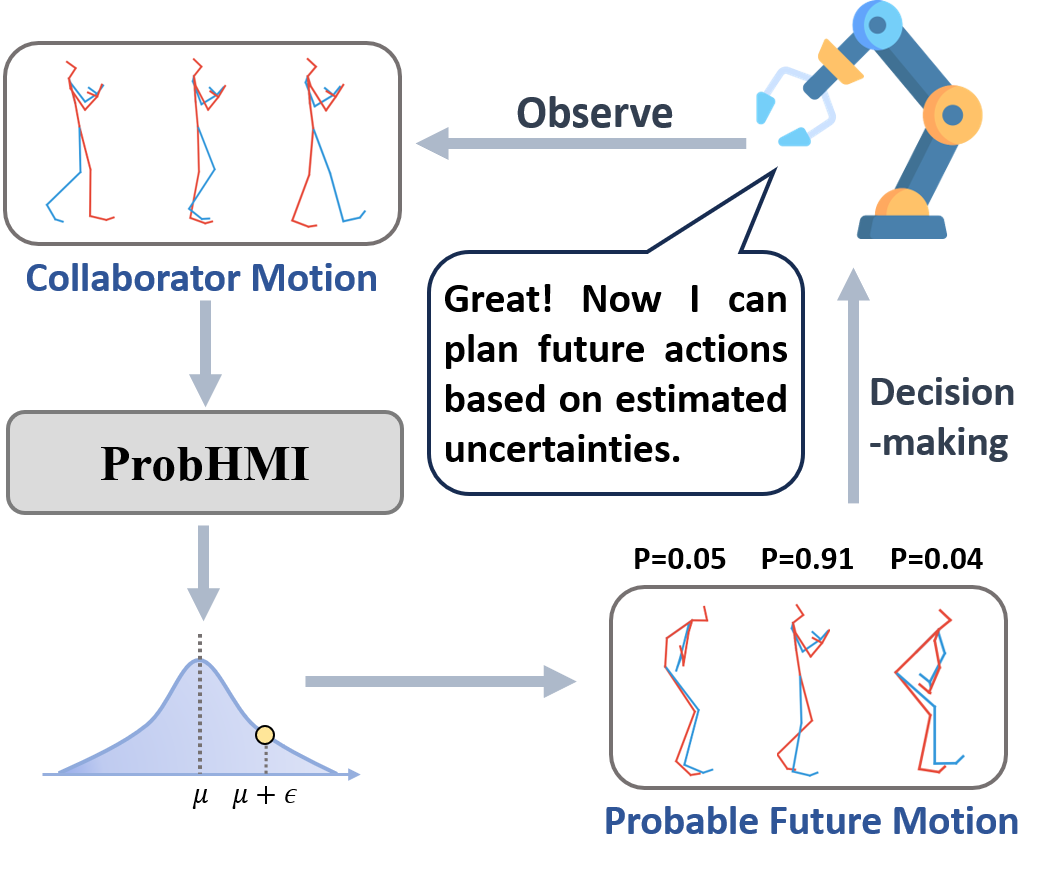}
            \label{fig:probabilistic}
        } \\
	\caption{\small Common diverse human motion forecasting (a) utilize the observation and codes drawn from Gaussian as the input of a generative model. Instead, our framework (b) explicitly models the distribution, enabling natural uncertainty quantification by probability density and quantile. The estimated uncertainty can then serve as a crucial basis for robots to plan their future actions, ensuring safe collaboration with humans.}
 \label{fig:comparasion_overview}
 \vspace{-0.2cm}
\end{figure}

Human motion forecasting involves anticipating 3D human motion from observed movements, which is crucial for ensuring safe human-robot collaboration (HRC). This allows robots to control and optimize their movements based on anticipated human motion \cite{el2021prednet,moon2021fast,kanazawa2019adaptive,sampieri2022pose,sun2024conformal}. Given the multi-modal and uncertain nature of human behavior, it is essential to generate a diverse, plausible and explainable distribution over possible future 3D motions to minimize risk and optimize decision \cite{kanazawa2019adaptive,sampieri2022pose,ren2023robots}. While recent works have made progress in improving forecast motion diversity using generative models \cite{yuan2020dlow,mao2021GSPS,zhang2021we,dang2022diverse,barsoum2018hp,g2017deligan,wei2023MotionDiff,chen2023humanmac}, two key limitations remain, as shown in \cref{fig:comparasion_overview}: 1) Without a probabilistic formulation, they cannot quantify prediction uncertainty, which is important for risk-aware control and planning \cite{sun2024conformal,romer2023vision}; 2) Sampling from implicit density models is inefficient, requiring many samples to accurately estimate the multi-modal motion distribution. 

Our \textbf{ProbHMI} presents a novel probabilistic framework to address the challenges of uncertainty quantification and diverse motion generation that existing methods cannot fully address. It formulates the problem by representing complex human poses in a continuous, disentangled latent space using invertible transformations, and predicting the future latent distribution on history. Specifically, we represent the prediction as a multi-dimensional Gaussian, where the mean corresponds to the ground truth in the dataset, and the variance measures the degree of diversity from the mean. This explicit probabilistic formulation enables the generation of diverse futures via sampling from forecast distribution, while allowing uncertainty to be quantified based on the likelihood -- an explicit measure lacking in traditional generative model-based methods. Here, diversity is defined as variations among the set of plausible ways a future motion could unfold, such as differences in speed, direction or overall trajectory, and uncertainty refers to the confidence associated with individual predicted motions. Additionally, the explicit probabilistic formulation allows for the use of various sampling methods beyond random sampling, enabling a limited number of samples to effectively cover the entire forecasting motion space. Although commonly used neural networks can also project high-dimensional data to  a lower-dimensional semantic space, the discontinuity of this mapped space often yield unnatural samples, limiting their effectiveness in ProbHMI compared to invertible networks.

We conduct quantitative and qualitative evaluations of our probabilistic motion forecasting framework on standard benchmarks, including the Human3.6M and HumanEva-\uppercase\expandafter{\romannumeral1} datasets. Even when utilizing only a single GRU layer to model motion dynamics within our framework, our approach achieves superior performance for both deterministic and diverse prediction scenarios. 
Additionally, we introduce an empirical quantile evaluation to validate the alignment between estimated uncertainty and actual outcomes.
Our main contributions are as follows:
\begin{itemize}
\item We introduce a novel probabilistic framework for 3D human motion forecasting that facilitates principled uncertainty quantification and efficient sampling.
\item We perform a series of experimental validations on benchmark datasets to assess our framework. The results do not only surpass baselines but also firmly establish the efficacy of our approach regarding uncertainty quantification and sampling efficiency. 
\end{itemize}

\section{Related Work}

\subsection{3D Human Motion Forecasting} 
Human motion forecasting has been widely studied using various techniques. Early works cast it as a regression task optimized by MSE loss within a recurrent encoder-decoder (RED) framework \cite{fragkiadaki2015recurrent,ghosh2017learning,martinez2017human,pavllo2020modeling,mao2020history}. While achieving high accuracy, these deterministic methods fail to represent the diverse nature of human motions. Subsequently, graph convolutional networks (GCNs) \cite{li2020dynamic,li2021multiscale,zhong2022spatio,chen2023spatiotemporal,zhang2024CHAMP} and transformers \cite{nargund2023spotr,ding2022towards,yu2023towards,tang2023collaborative} were explored to model temporal dynamics. In parallel, deep generative models including variational auto-encoders (VAEs) \cite{yuan2020dlow,mao2021GSPS,zhang2021we,dang2022diverse,ma2022multi,salzmann2022motron}, generative adversarial networks (GANs) \cite{barsoum2018hp,g2017deligan} and diffusion models \cite{wei2023MotionDiff,chen2023humanmac} were introduced to generate diverse futures. However, these methods incorporate independent sampling codes from standard Gaussian distributions as additional inputs, making predicted sequences indistinguishable with no correlation to uncertainty. 

\subsection{Uncertainty in Forecasting}
Uncertainty estimation has been a long-standing focus in time-series forecasting \cite{salinas2020deepar,stankeviciute2021conformal,toubeau2018deep,rasul2021autoregressive,sohl2015deep,wen1711multi}, particularly for high-stakes applications such as weather forecasting and stock price prediction. Typical approaches involve representing the future as a probability distribution, such as Gaussian distribution, and predicting its parameters, which allows for the use of probability density or quantile as uncertainty metrics. This paradigm is also widely applied in trajectory forecasting \cite{alahi2016social,zhao2021you,liang2020learning,mohamed2022social,salzmann2020trajectron++}, where human can be modeled as 2D point and represented with bi-variate Gaussian. However, directly extending this methodology to 3D human motion is challenging, since commonly used parametric distributions struggle to capture the complexity of 3D human motion, often resulting in unnatural predictions. In contrast, we introduce invertible networks to parameterize complex data distributions into a parametric form, enabling both explicit uncertainty estimation and plausible predictions.

\subsection{Invertible Networks} \label{SEC:InvertibleNetworks}

Invertible networks \cite{dinh2014nice,dinh2016density,kingma2018glow,behrmann2019invertible,chen2019residual} were initially designed as a form of deep probabilistic models, which consists of a series of bijective transformations to guarantee the invertibility, thus allowing for exact likelihood computation. 

Given an invertible transformation $f:\mathcal{X} \to \mathcal{Z}$ that maps a data distribution $P_\mathcal{X}$ to a simpler parametric distribution $P_\mathcal{Z}$, the inference from a  random variable $z$ following $P_\mathcal{Z}$ to the corresponding data $x$ is achieved by the inverse function $x = f^{-1}(z)$. As directly modeling $f$ is intractable, invertible networks employ a chain of simpler transformations $\{f_k\}^K_{k=1}$ to approximate $f$ as $f = f_1 \circ f_2 \circ \dots \circ f_K$.

Thus, we can represent the probability density of $x$ as:
   \begin{equation}
       P_\mathcal{X}(x) = P_\mathcal{Z}(z) \prod_{k = 1}^{K} |\det(\frac{\partial{z_k}}{\partial{z_{k-1}}})|,
   \label{EQ:PD}
   \end{equation}
\noindent where $z_k = f_k(f_{k-1}( \dots f_2(f_1(x))))$ and the determinant terms capture the volume change introduced by each transformation $f_k$ of the invertible network.

Then, the exact log-likelihood of $P_\mathcal{X}$ can be written as:
   \begin{equation}  
       \log{P_\mathcal{X}(x)} = \log{P_\mathcal{Z}(z)} + \sum_{k = 1}^{K}\log{|\det(\frac{\partial{z_k}}{\partial{z_{k-1}}})|}.
   \label{EQ:ML}
   \end{equation}

\begin{figure*}[htb]\captionsetup[subfloat]{font=small}
    \centering
    \subfloat[training phase] {
        \includegraphics[height=0.225\linewidth]
            {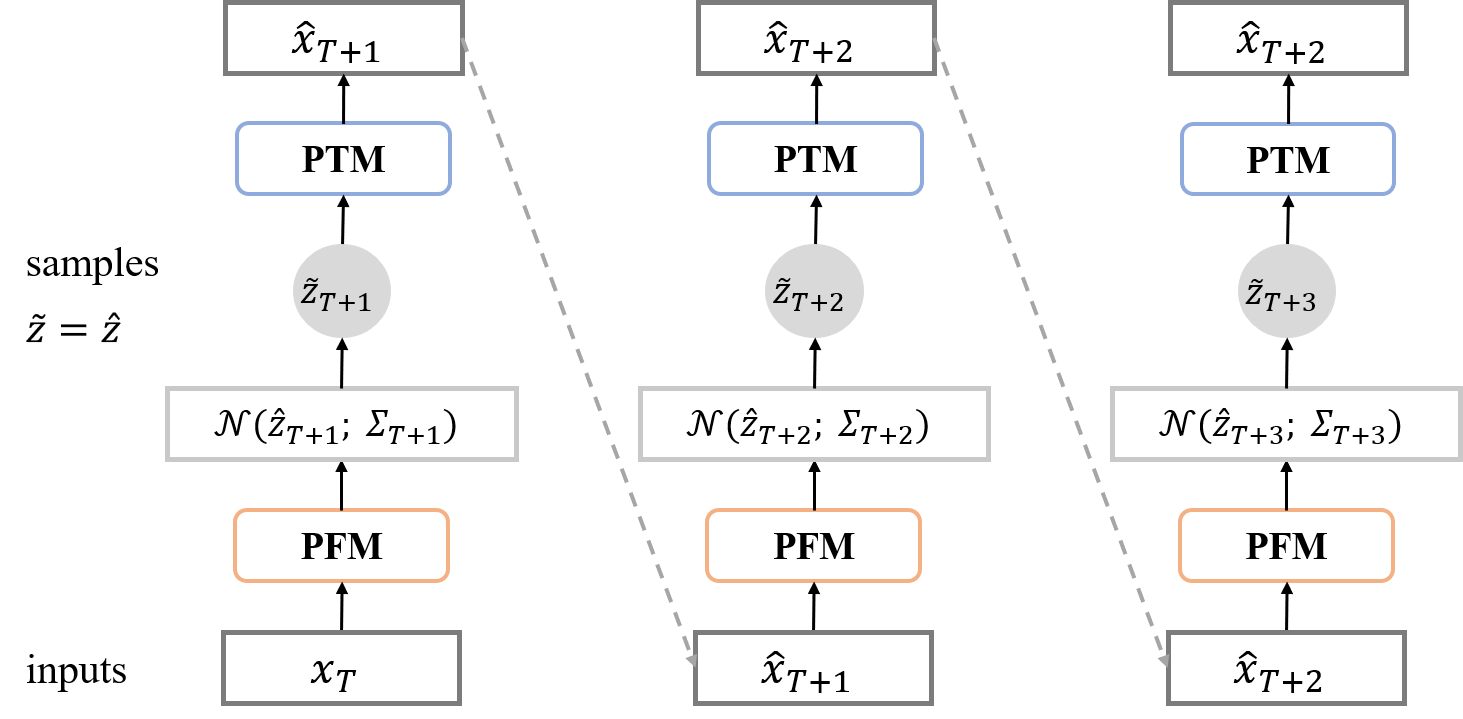}
        \label{fig:training}
    }
    \subfloat[inference phase]{
        \includegraphics[height=0.225\linewidth]
            {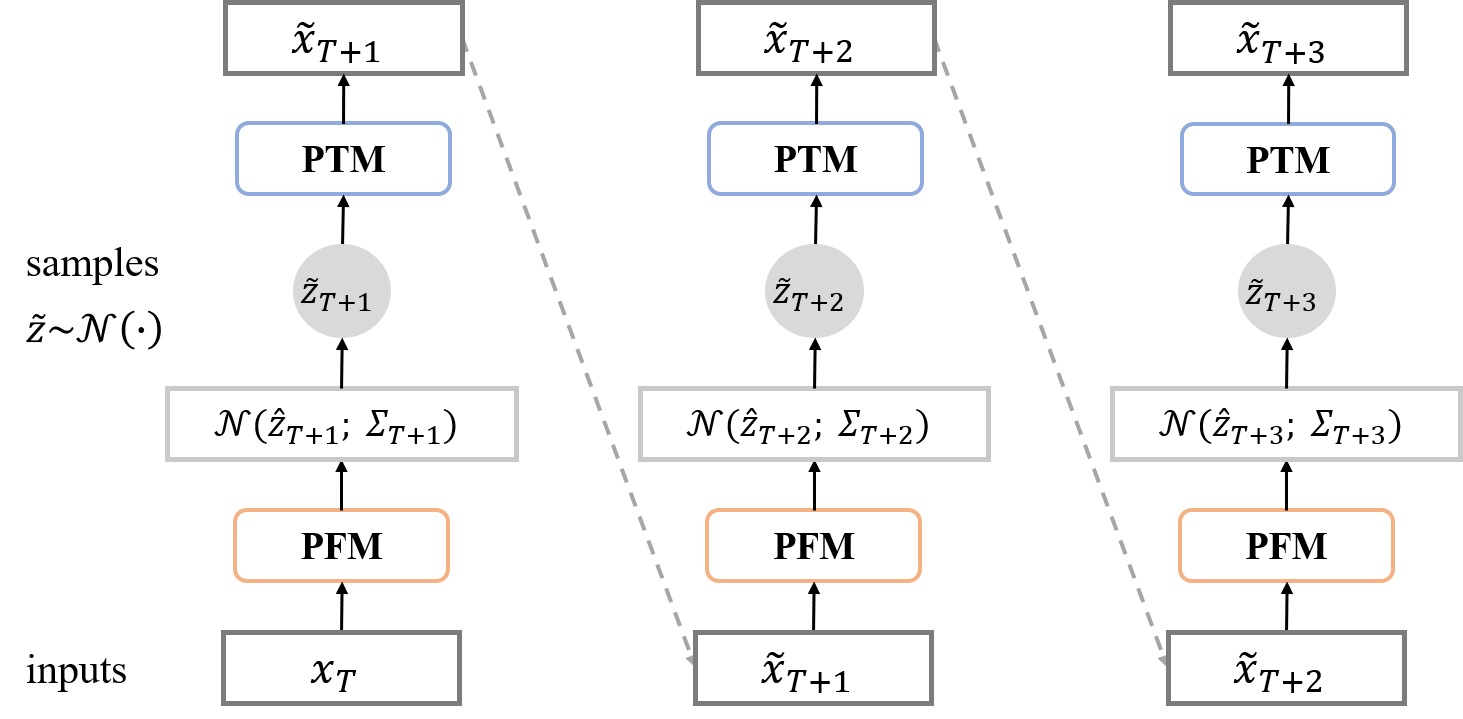}
        \label{fig:inference}
    }
    \caption{\small Overview of ProbHMI. During training, the PFM module uses the direct output from time $t$-1 to predict the distribution of the latent code at time $t$. At inference, multiple latent codes are drawn from the latent distribution, enabling diverse motion forecasting.}
    \label{fig:overview}
\end{figure*}

\section{Problem Formulation} \label{SEC:problem_formulation}
The goal of our approach is to predict a diverse set of 3D human motion while quantifying the associated uncertainty. We represent the input as a sequence of 3D human motion poses over $T$ frames, formally defined as $\mathbf{X}_{1:T} = \{\mathbf{x}_1, \mathbf{x}_2, \cdots, \mathbf{x}_{T}\}$. Here, $\mathbf{x}_t \in \mathbb{R}^{J \times C}$ represents a body pose with $J$ joints each containing $C$ channels at time $t$. Given $\mathbf{X}_{1:T}$, the direct output of our approach is the predicted pose sequence $\hat{\mathbf{X}}_{T+1:T+K} = \{\hat{\mathbf{x}}_{T+1}, \hat{\mathbf{x}}_{T+2}, \cdots, \hat{\mathbf{x}}_{T+K}\}$, which is comprising of $K$ frames and supervised by the corresponding ground truth $\mathbf{X}_{T+1:T+K}$. The diverse set of predictions is generated around $\hat{\mathbf{X}}_{T+1:T+K}$ and is defined as $\tilde{\mathbf{X}}_{T+1:T+K}^{1:S} ={\tilde{\mathbf{X}}_{T+1:T+K}^{1}, \tilde{\mathbf{X}}_{T+1:T+K}^{2}, \cdots, \tilde{\mathbf{X}}_{T+1:T+K}^{S}}$, where $S$ is the number of samples. Correspondingly, we define predicted latent codes associated with $\hat{\mathbf{X}}_{T+1:T+K}$ as $\hat{\mathbf{Z}}_{T+1:T+K} = \{\hat{\mathbf{z}}_{T+1}, \hat{\mathbf{z}}_{T+2}, \cdots, \hat{\mathbf{z}}_{T+K}\}$, and the latent codes corresponding to $\tilde{\mathbf{X}}_{T+1:T+K}^{1:S}$ as $\tilde{\mathbf{Z}}_{T+1:T+K}^{1:S} ={\tilde{\mathbf{Z}}_{T+1:T+K}^{1}, \tilde{\mathbf{Z}}_{T+1:T+K}^{2}, \cdots, \tilde{\mathbf{Z}}_{T+1:T+K}^{S}}$.

\section{Methodology}
In this section, we begin with the introduction of the  framework in \cref{SEC:framework_overview}, followed by detailed discussions of two key components respective in \cref{SEC:PTM} and \cref{SEC:PFM}. Then, we describe objective functions in \cref{SEC:LOSS}, and elaborate on the uncertainty quantification paradigm within ProbHMI in \cref{SEC:UQ}.

\subsection{Framework Overview} \label{SEC:framework_overview}
We propose ProbHMI (shown in \cref{fig:overview}), a probabilistic human motion forecasting framework consisting of two key components. The first component, Pose Transformation Module (PTM), is an invertible network that connects the latent space with the data space. The second component, Pose Forecasting Module (PFM) is responsible for forecasting future motions in the latent space built by the PTM module. 

In the training phase (shown in \cref{fig:training}), the PFM module takes the observation $\mathbf{X}_{1:T}$ and previous predicted results $\hat{\mathbf{X}}_{T+1:T+t}$ as input to forecast the conditional distribution $p(\hat{\mathbf{z}}_{T+t+1},\Sigma_{T+t+1})$. The PTM module then transforms $\hat{\mathbf{z}}_{T+t+1}$ to the corresponding pose $\hat{\mathbf{x}}_{T+t+1}$. This progress is repeated $K$ times to generate final sequence $\hat{\mathbf{X}}_{T+1:T+K}$. The parameters of ProbHMI can be optimized by minimizing the negative log-likelihood between the predicted distribution and the actual latent codes $\mathbf{Z}_{T+1:T+K}$, which are transformed from $\mathbf{X}_{T+1:T+K}$ by the PTM module.

In the inference phase (shown in \cref{fig:inference}), the primary difference from the training phase is the inclusion of a sampling process that draws $\tilde{\mathbf{z}}_{T+t+1}$ from $p(\hat{\mathbf{z}}_{T+t+1},\Sigma_{T+t+1})$. The corresponding $\tilde{\mathbf{x}}_{T+t+1}$ becomes not only the forecasting result but also the input for the next iteration. By repeatedly sampling, ProbHMI can generate multiple diverse future motion sequences.

In summary, the dynamics of ProbHMI is formulated as \cref{EQ:dynamics}, where the dynamics of the training phase can be seen as a special case of the inference phase.
\begin{equation}
    \begin{split}
    \hat{\mathbf{z}}_{T+t+1}, \Sigma_{T+t+1} & = {\rm PFM}(\mathbf{X}_{1:T}, \tilde{\mathbf{x}}_{T+1:T+t}) \\
    \epsilon_{T+t+1} & \sim \beta_{T+t+1} \mathcal{N}(0, \Sigma_{T+t+1}) \\
    \tilde{\mathbf{z}}_{T+t+1} & = \hat{\mathbf{z}}_{T+t+1} + \epsilon_{T+t+1} \\ 
    \tilde{\mathbf{x}}_{T+t+1} & = {\rm PTM}(\tilde{\mathbf{z}}_{T+t+1}),
    \end{split}
\label{EQ:dynamics}
\end{equation}
where $\beta$ controls the practical variance during sampling, and is set to $0$ in the training phase.

\subsection{Pose Transformation Module}
\label{SEC:PTM}
The PTM module is responsible for transforming motion representations between human skeleton poses and latent codes, followed by the foundation of invertible networks as described in \cref{SEC:InvertibleNetworks}. However, standard invertible networks can disrupt these structural relationships during transformation since human skeletal poses are inherently structured with spatial dependencies between joints, while standard invertible networks perform channel-wise operations. Thus, we introduce a part-aware invertible network based on NICE \cite{dinh2014nice}, which preserves the topological structure throughout the transformation, leveraging the inherent graph-based structure of the human skeleton by designing transformations to operate on hierarchical body parts such as joints, limbs, and the full body.  

Specifically, we introduce a \emph{GCN-based additive coupling layer} which is formulated as \cref{EQ:gacl}.
    \begin{equation}
        \begin{split}
            H^{l+1}_{I_1} & = H^{l}_{I_1}, \\ 
            H^{l+1}_{I_2} & = H^{l}_{I_2} + AH^{l}_{I_1}W^{l},
        \end{split}
    \label{EQ:gacl}
    \end{equation}
where $H^{l}$ denoting the feature graph produced by the $l$-th layer, and $(H^{l}_{I_1}, H^{l}_{I_2})$ represent the graph partitions of $H^{l}$ based on human topology, such as $H^{l}_{I_1}$ and $H^{l}_{I_2}$ representing the upper body and the lower body, respectively. The adjacency matrix $A\in\mathbb{R}^{N\times N}$ encodes the skeletal connections between each node, where $N$ is the number of nodes in $H^{l}_{I_1}$. $W^l\in\mathbb{R}^{F_{in}\times F_{out}}$ is the trainable transformation matrix, where $F_{in}$ and $F_{out}$ represent the dimensions of the input and output features of each node, respectively.

\subsection{Pose Forecasting Module} \label{SEC:PFM}
The PFM module models the motion dynamics over time through a structure not restricted to any single model. To demonstrate ProbHMI effectively, our implementation employs a single GRU layer - a simple recurrent network capable of learning sequences. While other recurrent architectures could also potentially capture temporal dependencies, the GRU sufficed here to validate the framework.


Given the complexity of modeling whole-body motion, directly predicting dynamics across all joints can be challenging. As the human skeletal system consists of interconnected but semi-independent parts that move in coordinated yet distinct patterns, we introduce a part-aware prediction paradigm that predicts the future states of different body parts separately and in parallel. By accommodating the unique movement patterns of each part, this paradigm enhances both the accuracy and diversity in predictions.
    
\subsection{Loss Functions} \label{SEC:LOSS}
We train ProbHMI end-to-end in both the latent space and the pose space. $L_H$, the objective to maximize the likelihood of the predicted distribution within the latent space, is described as \cref{EQ:loss1}:
    \begin{equation}
        L_{H} = \frac{1}{K} \sum_{i=T+1}^{T+K} (\log(\Sigma_{i}) + \frac{(\mathbf{z}_{i} - \hat{\mathbf{z}}_{i})^2}{2(\Sigma_{i})^2}),
    \label{EQ:loss1}
    \end{equation}
\noindent where $\mathbf{z}$ is mapping from the PTM module and $\hat{\mathbf{z}}$ and $\Sigma$ denote the mean and the variance of the factorized Gaussian predicted by the PFM module, respectively.


$L_R$, the objective defined in the pose space, aims to minimize the $L1$ distance between predictions and the ground truth. It is expressed as \cref{EQ:reconstruct}:
    \begin{equation}
        L_{R} = \parallel \mathbf{X}_{T+1:T+K} -  \hat{\mathbf{X}}_{T+1:T+K} \parallel_1.
    \label{EQ:reconstruct}
    \end{equation}

Besides $L_H$ and $L_R$, we introduce $L_N$ as a regularization term for the PTM module, aiming to minimize the KL Divergence between the predicted distribution $p_\theta(\mathbf{\hat{Z}})$ and standard factorized Gaussian distribution. Supposed that $g(\mathbf{Z}) = \mathcal{N}(\mathbf{Z} | 0, I)$, it can be formulated as:
    \begin{equation}
        L_{N} = -\log g(\hat{\mathbf{Z}}) - \log{|\det(\frac{\partial{f^{-1}}}{\partial{\hat{\mathbf{X}}}})|}.
    \label{EQ:normal-likelihood}
    \end{equation}

In summary, the objective function of our approach is:
    \begin{equation}
        L = \alpha L_{H} + \beta L_{R} + \gamma L_{N},
    \label{EQ:totalloss}
    \end{equation}
where $\alpha, \beta, \gamma$ are the corresponding coefficients, and set to 0.1, 1.0, and 5.0, respectively.

\subsection{Uncertainty Quantification} \label{SEC:UQ}

\subsubsection{Frame-level Uncertainty}
We represent the frame's uncertainty as the quantile associated with the latent code for this pose, which is straightforward to calculate.

\subsubsection{Sequence-level Uncertainty}
Since our method is an auto-aggressive model, direct comparison among sampled sequences may be unclear. To ensure a meaningful uncertainty quantification, we apply the same quantile to all frames of the sequence during sampling, and use this quantile to represent the uncertainty of the sequence. We follow this sampling schedule in our experiments.

\section{Experiments}
In this section, we first measure the predictive performance of ProbHMI with respect to baselines in \cref{SEC:Diverse_Eval} and \cref{SEC:Deterministic_Eval}. Second, we validate the proposed uncertainty quantification paradigm and the sampling efficiency of ProbHMI in \cref{SEC:EUQ} and \cref{SEC:EES}, respectively.
Finally, we provide an ablation study to validate the benefit of introduced part-aware paradigms in \cref{EQ:Ablation}.

\subsection{Datasets}
\subsubsection{Human3.6M}
We evaluate ProbHMI on Human3.6M \cite{ionescu2013human3} for both diverse and deterministic setups. In the diverse setup, we utilize 25 observed frames (0.5s) followed by 100 prediction frames (2s) at 50 fps, with a 17-joint skeleton. Following the previous work \cite{yuan2020dlow}, we train on (S1, S5, S6, S7, S8) and test on (S9, S11). In the deterministic setup, we utilize 10 observed frames (0.4s) followed by 25 prediction frames (1s) which are down-sampled to 25 fps. We adopt a 22-joint skeleton following \cite{salzmann2022motron} and train on (S1, S6, S7, S8, S9, S11) and test on S5. We represent human pose by exponential maps on Human3.6M.

\subsubsection{HumanEva-\uppercase\expandafter{\romannumeral1}}
We evaluate ProbHMI on HumanEva-\uppercase\expandafter{\romannumeral1} \cite{sigal2010humaneva} on the diverse setup. We utilize 15 observed frames (0.25s) followed by 60 prediction frames (1s) at 60 fps. The split of the dataset follows the official set. As HumanEva-\uppercase\expandafter{\romannumeral1} does not include joint angles, we represent human pose using Cartesian coordinates.

\subsection{Diverse Evaluation} \label{SEC:Diverse_Eval}

\subsubsection{Metrics}
Following prior works \cite{yuan2020dlow,mao2021GSPS,zhang2021we}, we use six evaluation metrics including \textbf{APD} (Average Pairwise Distance), \textbf{ADE} (Average Displacement Error), \textbf{FDE} (Final Displacement Error), \textbf{MMADE} (Multi-Modal-ADE), \textbf{MMFDE} (Multi-Modal-FDE) and \textbf{FID} (Fr\'echet Inception Distance), with all metrics computed based on 50 samples.

\subsubsection{Diverse Baselines}
We compare ProbHMI with following baselines: (1) GAN-based methods (\textbf{HP-GAN} \cite{barsoum2018hp}, \textbf{DeLiGAN} \cite{g2017deligan}). (2) VAE-based methods (\textbf{BoM} \cite{b2018accurate}, \textbf{DLow} \cite{yuan2020dlow}, \textbf{GSPS} \cite{mao2021GSPS}, \textbf{MOJO} \cite{zhang2021we}, \textbf{DivSamp} \cite{dang2022diverse}, \textbf{Motron} \cite{salzmann2022motron}, \textbf{MutltiObj} \cite{ma2022multi}). (3) Diffusion-based methods (\textbf{MotionDiff} \cite{wei2023MotionDiff}, \textbf{HumanMAC} \cite{chen2023humanmac}).

\begin{table}[htb]
    \renewcommand{\arraystretch}{1.25}
    \centering
    \small
    \caption{\small The diverse evaluation results on Human3.6M.}
    \resizebox{1.0\linewidth}{!}{
        \begin{tabular}{lr|cccccc}
        \toprule
             & Params & APD$\uparrow$ & ADE$\downarrow$ & FDE$\downarrow$ & MMADE$\downarrow$ & MMFDE$\downarrow$ & FID$\downarrow$ \\
        \hline
             DLow & 7.30M & 11.741 & 0.425 & 0.518 & 0.495 & 0.531 & 1.255\\
             GSPS & 1.31M & 14.757 & 0.389 & 0.496 & \textbf{0.476} & 0.525 & 2.103\\
             MOJO & - & 12.579 & 0.412 & 0.514 & 0.497 & 0.538 & - \\
             DivSamp & 21.33M & 15.310 & 0.370 & 0.485 & 0.477 & \textbf{0.516} & 2.083 \\
             MultiObj & - & 14.240 & 0.414 & 0.516 & - & - & - \\
             Motron & 1.67M & 7.168 & 0.375 & 0.488 & 0.509 & 0.539 & 13.743\\
        \hline
             MotionDiff & 29.93M & \textbf{15.353} & 0.411 & 0.509 & 0.508 & 0.536 & - \\
             HumanMAC & 28.40M & 6.301 & 0.369 & \textbf{0.480} & 0.509 & 0.545 & - \\
        \hline
             ProbHMI & \textbf{0.36M} & 6.682 & \textbf{0.364} & 0.493 & 0.511 & 0.558 & \textbf{0.646} \\
        \bottomrule
        \end{tabular}
        \label{TAB:diverse_quan_result}
    }
\end{table}

\begin{table}[htb]
    \renewcommand{\arraystretch}{1.25}
    \centering
    \caption{\small The diverse evaluation results on HumanEva-\uppercase\expandafter{\romannumeral1}.}
    \resizebox{0.85\linewidth}{!}{
        \begin{tabular}{l|ccccc}
        \toprule
            & APD$\uparrow$ & ADE$\downarrow$ & FDE$\downarrow$ & MMADE$\downarrow$ & MMFDE$\downarrow$ \\
        \hline
            HP-GAN   & 1.139 & 0.772 & 0.749 & 0.776 & 0.769 \\
            DeLiGAN  & 2.177 & 0.306 & 0.322 & 0.385 & 0.371 \\
            BoM  & 2.846 & 0.271 & 0.279 & 0.373 & 0.351 \\
            DLow & 4.855 & 0.251 & 0.268 & 0.362 & 0.339 \\
            GSPS & 5.825 & 0.233 & 0.244 & \textbf{0.343} & 0.331 \\
            MOJO & 4.181 & 0.234 & 0.244 & 0.369 & 0.347 \\
            MotionDiff & \textbf{5.931} & 0.232 & \textbf{0.236} & 0.352 & \textbf{0.320} \\
        \hline
            ProbHMI & 4.810 & \textbf{0.211} & 0.245 & 0.416 & 0.418 \\
        \bottomrule
        \end{tabular}
        \label{TAB:humaneva_results}
    }
\end{table}

\subsubsection{Quantitative results} The results of ProbHMI against diverse approaches are presented in \cref{TAB:diverse_quan_result} (on Human3.6M) and \cref{TAB:humaneva_results} (on HumanEva-\uppercase\expandafter{\romannumeral1}). It demonstrates that ProbHMI achieves superior performance to prior methods, particularly on ADE and FID. As ProbHMI autoregressively forecasts future poses that draws each subsequent pose from a distribution based on the previous time step, the value of FDE is slightly higher than other methods, which reflects the real-world principle that uncertainty increases over time.

We also report the average computational time in \cref{TAB:comp_time_average}. ProbHMI achieves real-time prediction and performs much faster than Motron and HumanMAC. The reason for the slower performance compared to DLow is that separate parts within ProbHMI must be computed sequentially in PyTorch.

\begin{table}[htb]
    \renewcommand{\arraystretch}{1.25}
    \centering
    \caption{\small The average inference time on Human3.6M. All results were conducted on a  NVIDIA 2080Ti GPU and a Intel(R) Xeon(R) Gold 5120T CPU, and represent the mean of 1000 tests.}
    \resizebox{0.9\linewidth}{!}{
        \begin{tabular}{c|cccc}
        \toprule
            Dataset & ProbHMI & DLow & Motron  & HumanMAC \\
        \hline
            \textbf{Human3.6M} & 195ms & 95ms & 475ms & 3453ms \\
        \bottomrule
        \end{tabular}
        \label{TAB:comp_time_average}
    }
\end{table}

\subsubsection{Qualitative results}
\begin{figure*}[htb]
  \centering
  \resizebox{1.0\linewidth}{!}{
    \begin{tabular}{cc} 
        \vspace{5.0em}
        
        \begin{tabular}{c} \huge
            \rotatebox[origin=c]{90}{Human3.6M}
        \end{tabular} &
        \begin{tabular}{cccc}
            \begin{tabular}{l}
                \subfloat{\includegraphics[width=1.0\linewidth]{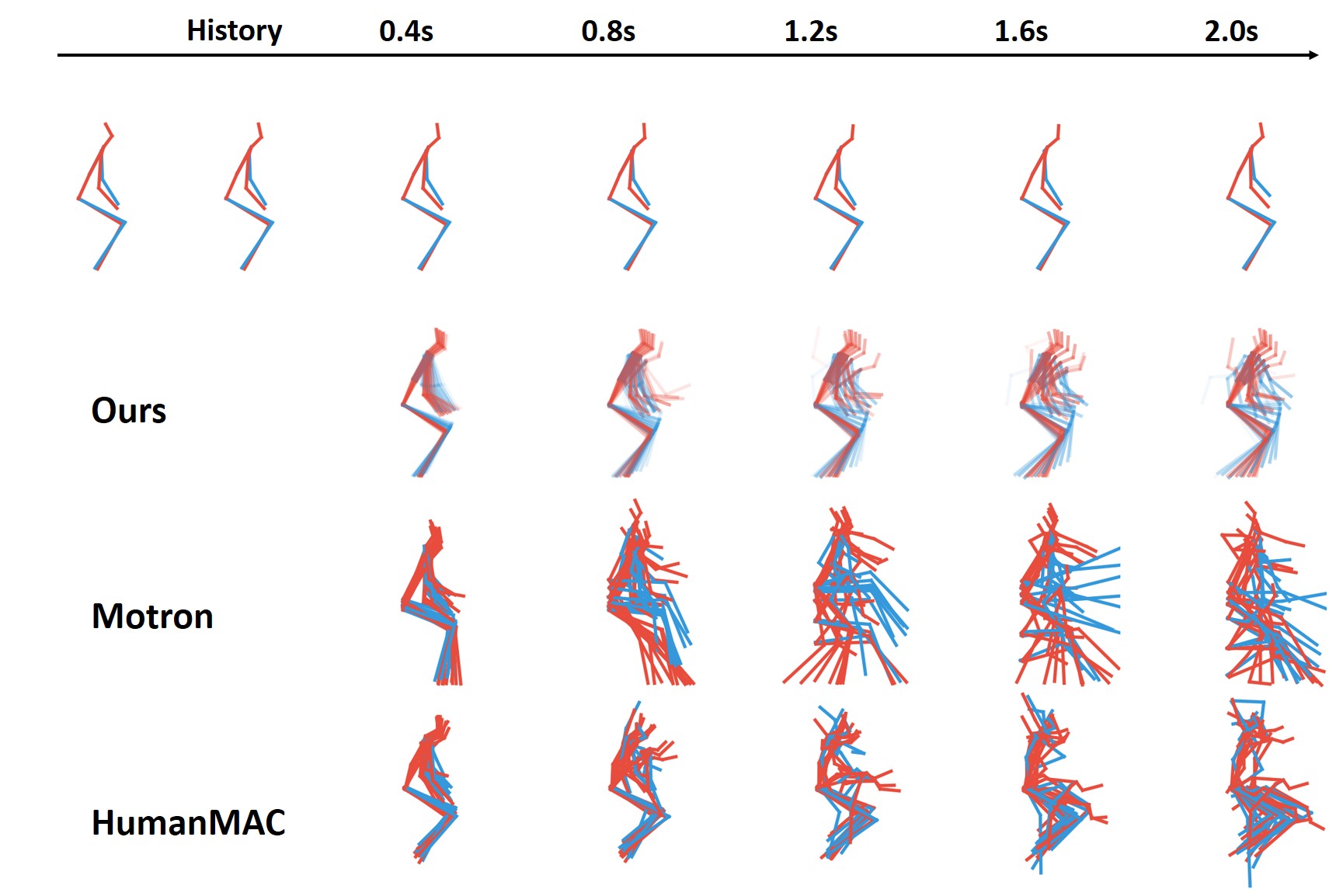}} 
            \end{tabular} &
            
            \begin{tabular}{l}
                \subfloat{\includegraphics[width=1.0\linewidth]{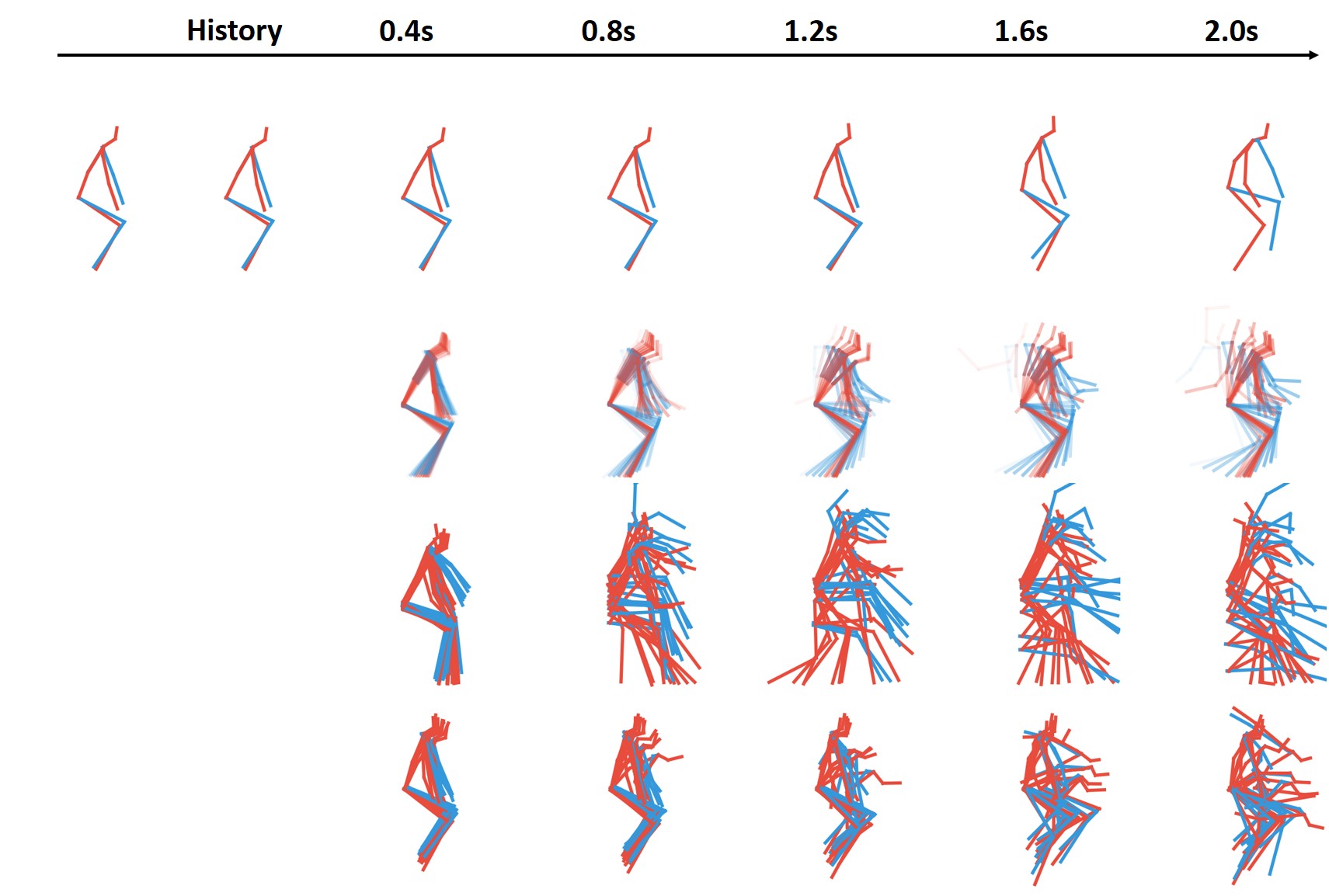}} 
            \end{tabular} &
    
            \begin{tabular}{l}
                \subfloat{\includegraphics[width=1.0\linewidth]{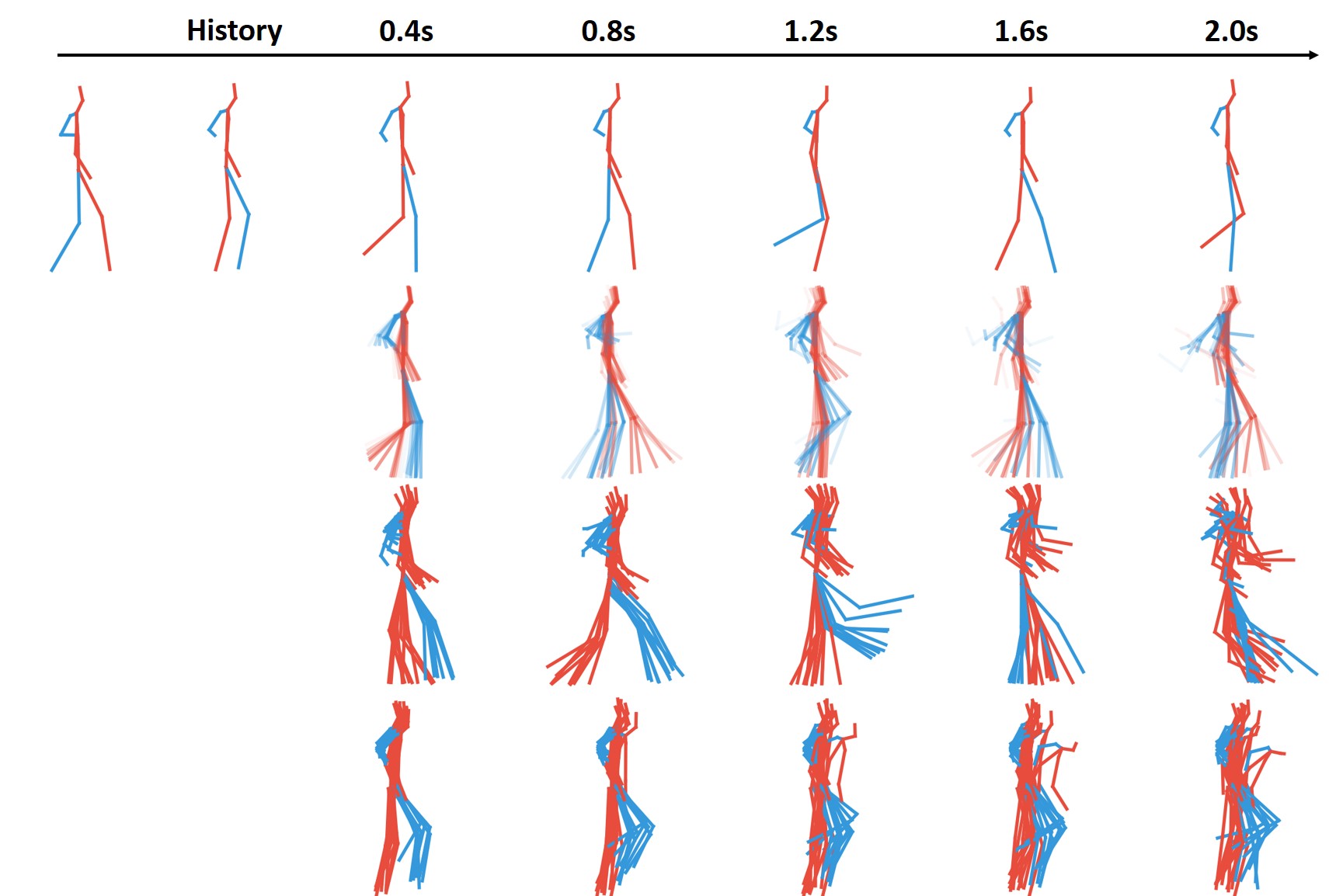}} 
            \end{tabular} &
            
            \begin{tabular}{l}
                \subfloat{\includegraphics[width=1.0\linewidth]{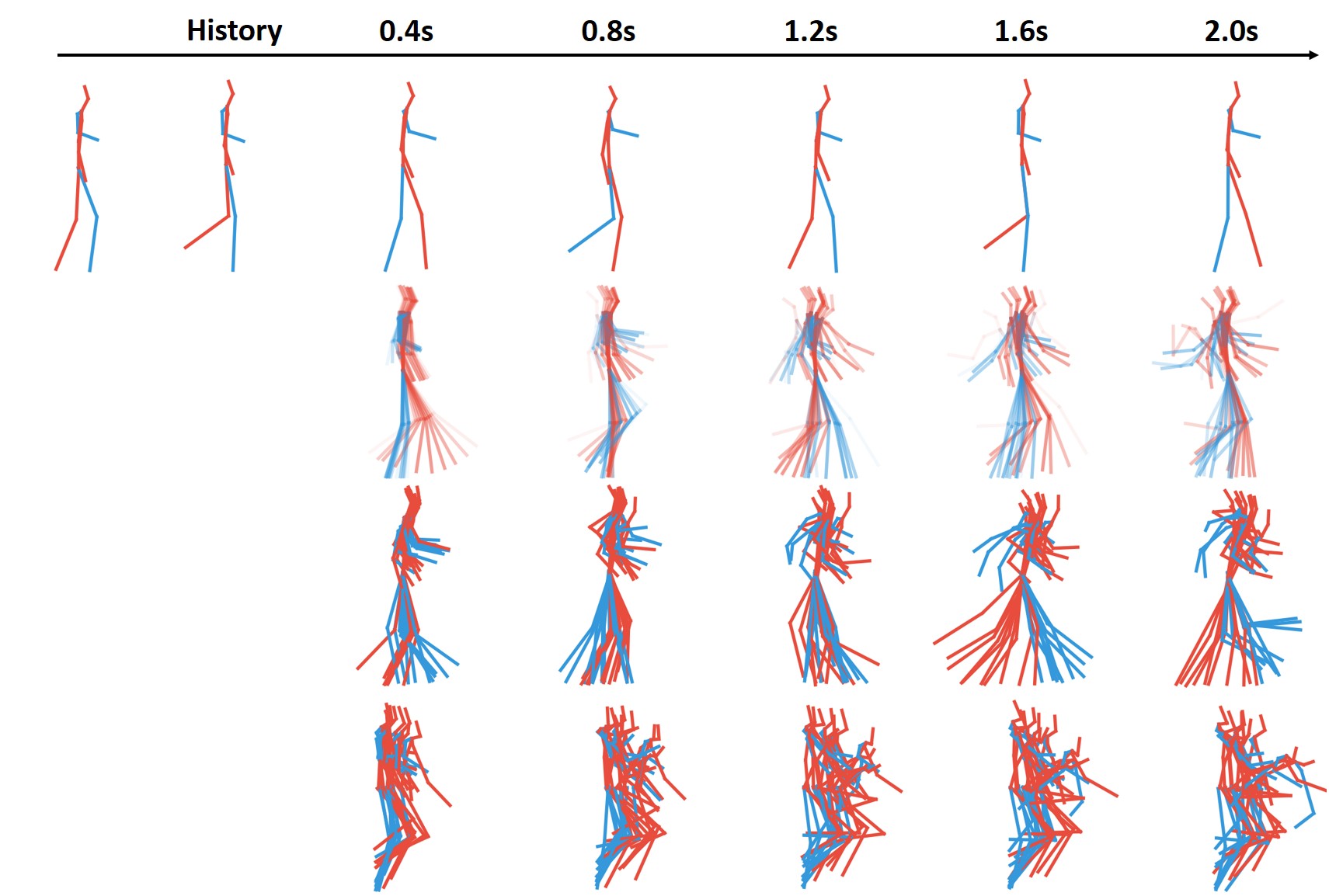}} 
            \end{tabular}
        \end{tabular} \\
        
        \begin{tabular}{c} \huge
            \rotatebox[origin=c]{90}{HumanEva-\uppercase\expandafter{\romannumeral1}}
        \end{tabular} &
        \begin{tabular}{cccc}
            \begin{tabular}{l}
                \subfloat{\includegraphics[width=1.0\linewidth]{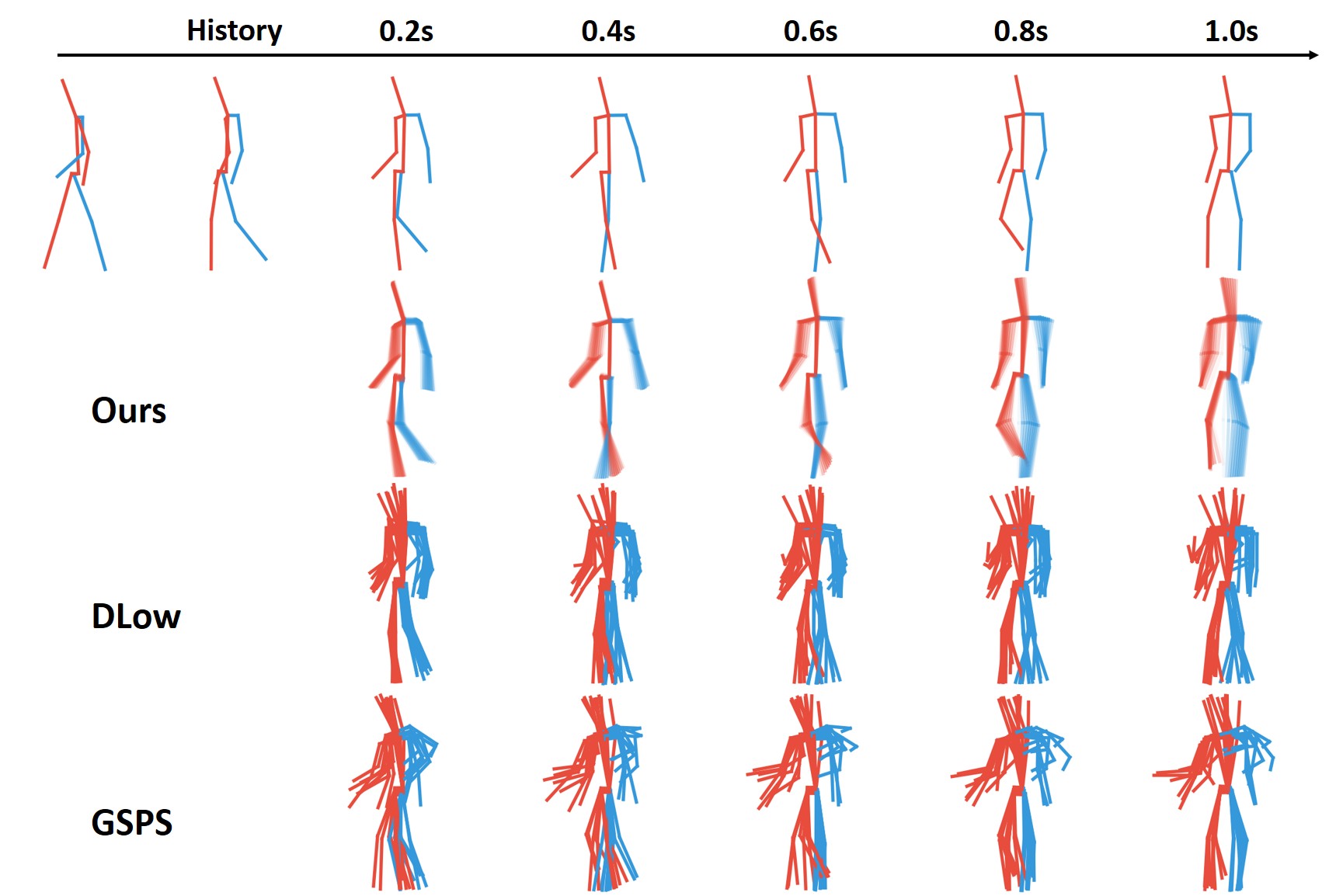}} 
            \end{tabular} &
            
            \begin{tabular}{l}
                \subfloat{\includegraphics[width=1.0\linewidth]{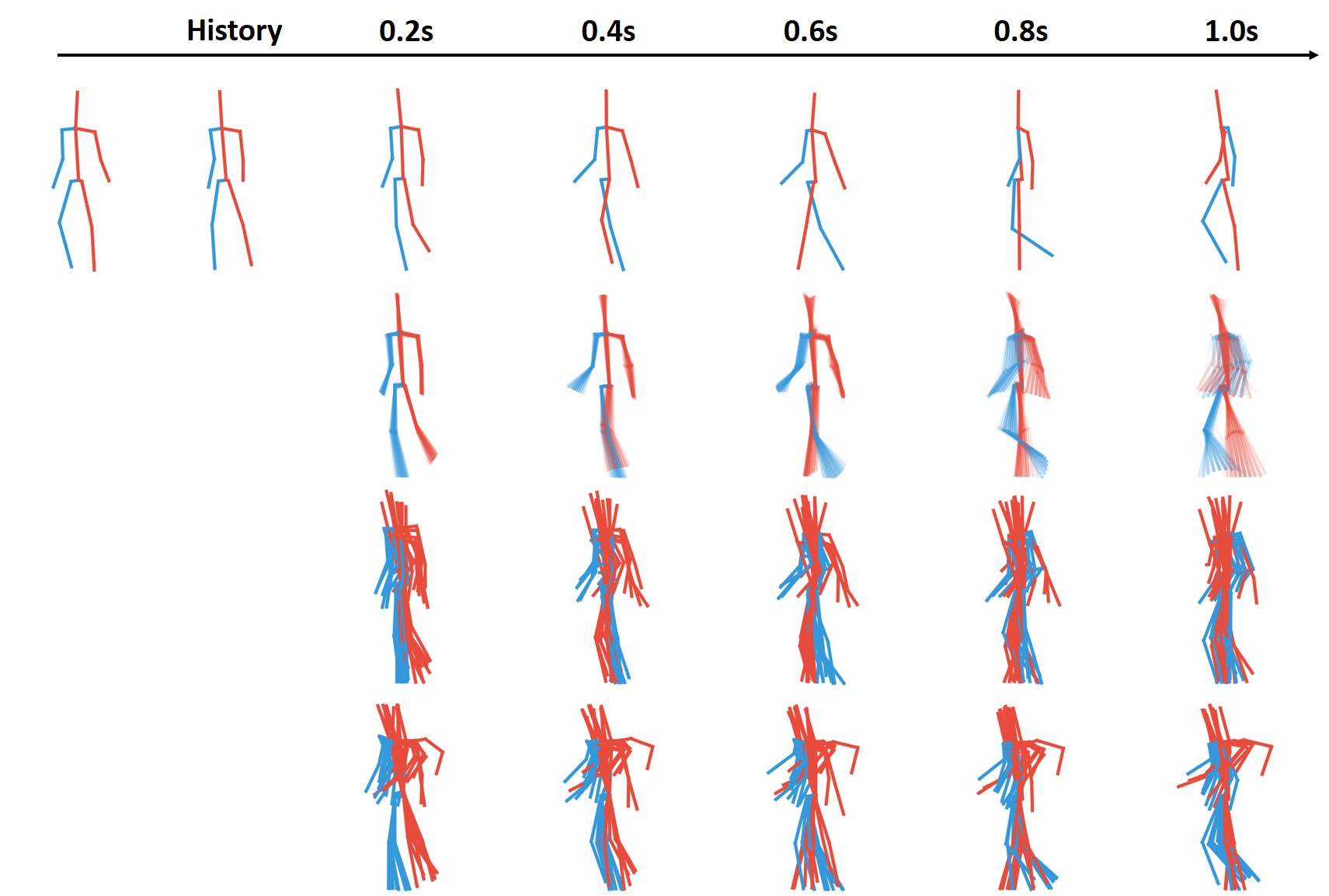}} 
            \end{tabular} &
    
            \begin{tabular}{l}
                \subfloat{\includegraphics[width=1.0\linewidth]{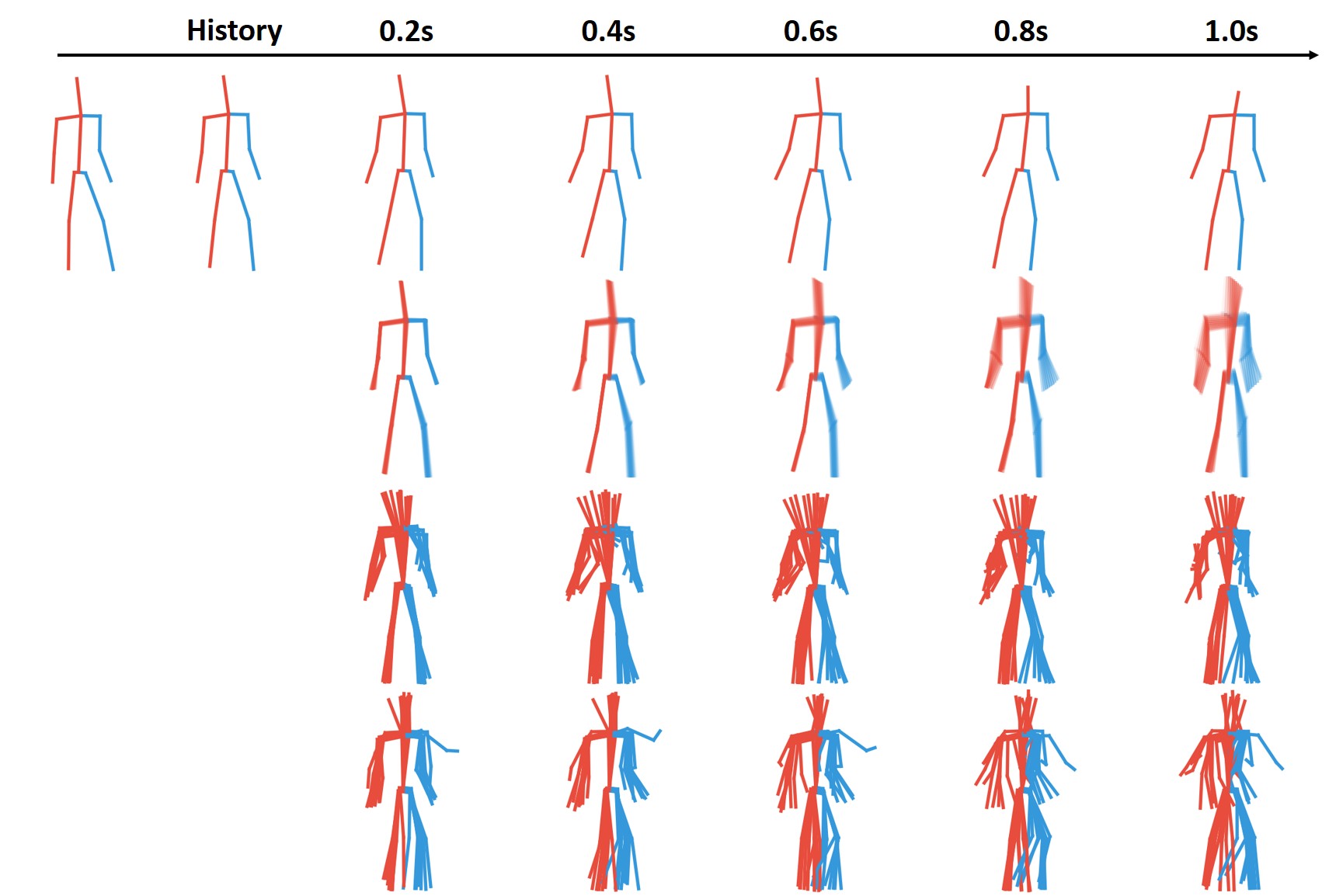}} 
            \end{tabular} &
            
            \begin{tabular}{l}
                \subfloat{\includegraphics[width=1.0\linewidth]{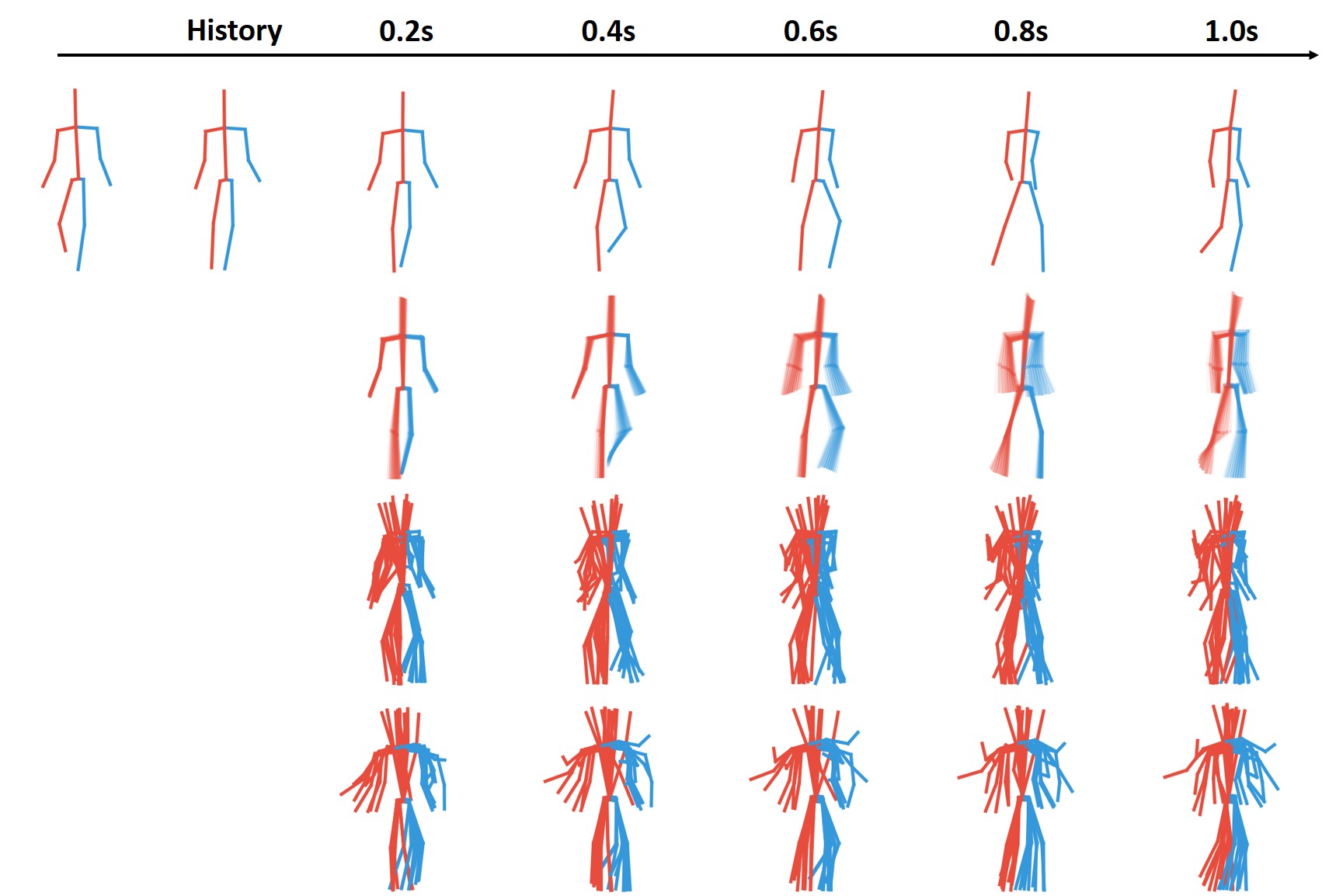}} 
            \end{tabular}
        \end{tabular}

    \end{tabular}
  }
\caption{\small Visualization results. We present qualitative comparison results with Motron \cite{salzmann2022motron} and HumanMAC \cite{chen2023humanmac} on the Human3.6M dataset (top), and with DLow \cite{yuan2020dlow} and GSPS \cite{mao2021GSPS} on the HumanEva-\uppercase\expandafter{\romannumeral1} dataset (bottom). The results of our method are weighted by quantile as estimated by ProbHMI, with greater opacity indicating higher quantile.} 
\label{fig:qualitative_results}
\end{figure*}

We present the visualization of predictions in comparison in \cref{fig:qualitative_results}, where each result consists of 10 samples. Although some baselines are capable of generating diverse motions, their results include many failure cases characterized by a large distance from the ground truth and a lack of reasonableness. In contrast, the predicted poses generated by ProbHMI remain centered around the ground truth, even after multiple samplings, demonstrating much higher fidelity. Moreover, It also demonstrates the effectiveness of our method in producing diverse predictions with quantified uncertainty, where the
predicted pose with higher quantile aligns more closely with the ground truth.


\subsection{Deterministic Evaluation} \label{SEC:Deterministic_Eval}
\subsubsection{Metrics}
We evaluate the \textbf{Mean Angle Error (MAE)} on the angle space, calculated as the average L2 distance across all angles between the predicted sequence and ground truth, following \cite{martinez2017human}.

\subsubsection{Deterministic Baselines}
We compare ProbHMI with deterministic approaches that use joint angles as the pose representation, including \textbf{ResGRU} \cite{martinez2017human}, \textbf{DMGNN} \cite{li2020dynamic}, \textbf{Hisrep} \cite{mao2020history} and \textbf{Motron} \cite{salzmann2022motron}.

\begin{table}[htb]
    \renewcommand{\arraystretch}{1.25}
    \centering
    \small
    \caption{\small The deterministic evaluation results on Human3.6M.}
    \resizebox{1.0\linewidth}{!}{
        \begin{tabular}{lc|cccccc}
        \toprule
             & Params & 80ms & 160ms & 320ms & 400ms & 560ms & 1000ms \\
        \hline
            ResGRU & 3.44M & 0.40 & 0.69 & 1.04 & 1.18 & - & - \\
            DMGNN & 62M & 0.27 & 0.52 & 0.83 & 0.95 & 1.17 & 1.57 \\
            Hisrep & 3.24M & 0.27 & 0.52 & 0.82 & 0.93 & 1.14  & 1.59 \\
            Motron & 1.67M & 0.26 & 0.48 & 0.82 & 0.95 & 1.15 & 1.60 \\
        \hline
            ProbHMI & \textbf{0.31M} & \textbf{0.26} & \textbf{0.46} & \textbf{0.73} & \textbf{0.86} & \textbf{1.11} & \textbf{1.48} \\
        \bottomrule
        \end{tabular}
        \label{TAB:deterministic_quan_result}
    }
\end{table}

\subsubsection{Quantitative Results} 
We report average MAEs across all actions in \cref{TAB:deterministic_quan_result}. For a fair comparison, we use only $\hat{\mathbf{X}}_{T+1:T+K}$ in the deterministic evaluation. Compared with prior works, ProbHMI achieves superior performances both in short-term prediction ($\le$ 400ms) and in long-term prediction ($\ge$ 500ms). Notably, as ResGRU employs a similar architecture to ProbHMI---minus the PTM module---ProbHMI's superior results \textbf{(0.86 vs. 1.18)} can highlight the effectiveness of forecasting in the latent space constructed by invertible networks compared to the original pose space. 

\begin{figure*}[htb]
  \centering
  \resizebox{0.97\linewidth}{!}{
    \begin{tabular}{cccc} 
        \begin{tabular}{c} \normalsize
        \end{tabular} &
        \begin{tabular}{ccccccc}
            \hspace{6.0em}
            \begin{tabular}{c} \huge
              \rotatebox[origin=c]{0}{History}
            \end{tabular} &
            \hspace{1em}
            \begin{tabular}{c} \huge
              \rotatebox[origin=c]{0}{0.2S}
            \end{tabular} &
            \hspace{1em}
            \begin{tabular}{c} \huge
              \rotatebox[origin=c]{0}{0.4S}
            \end{tabular} &
            \hspace{1em}
            \begin{tabular}{c} \huge
              \rotatebox[origin=c]{0}{0.8S}
            \end{tabular} &
            \hspace{2em}
            \begin{tabular}{c} \huge
              \rotatebox[origin=c]{0}{1.2S}
            \end{tabular} &
            \hspace{2em}
            \begin{tabular}{c} \huge
              \rotatebox[origin=c]{0}{1.6S}
            \end{tabular} &
            \hspace{2em}
            \begin{tabular}{c} \huge
              \rotatebox[origin=c]{0}{2.0S}
            \end{tabular}
        \end{tabular} &
        \begin{tabular}{c} \normalsize
        \end{tabular} &
        \begin{tabular}{ccccccc}
            \hspace{6.0em}
            \begin{tabular}{c} \huge
              \rotatebox[origin=c]{0}{History}
            \end{tabular} &
            \hspace{1em}
            \begin{tabular}{c} \huge
              \rotatebox[origin=c]{0}{0.2S}
            \end{tabular} &
            \hspace{1em}
            \begin{tabular}{c} \huge
              \rotatebox[origin=c]{0}{0.4S}
            \end{tabular} &
            \hspace{1em}
            \begin{tabular}{c} \huge
              \rotatebox[origin=c]{0}{0.8S}
            \end{tabular} &
            \hspace{2em}
            \begin{tabular}{c} \huge
              \rotatebox[origin=c]{0}{1.2S}
            \end{tabular} &
            \hspace{2em}
            \begin{tabular}{c} \huge
              \rotatebox[origin=c]{0}{1.6S}
            \end{tabular} &
            \hspace{2.0em}
            \begin{tabular}{c} \huge
              \rotatebox[origin=c]{0}{2.0S}
            \end{tabular}
        \end{tabular} \\

        \vspace{-0.2em}
        \begin{tabular}{c}
          \rotatebox[origin=c]{0}{\huge P50} \\
          \rotatebox[origin=c]{0}{\huge GT}
        \end{tabular} & \hspace{-2em}
        \begin{tabular}{l}
            \subfloat{\includegraphics[width=1.3\linewidth]{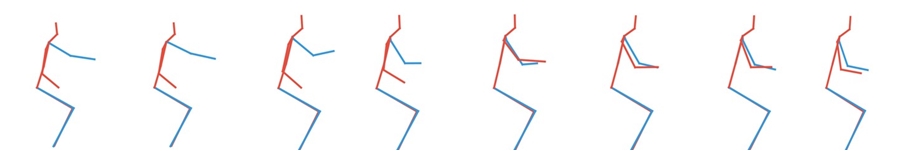}} 
        \end{tabular} & \hspace{2.5em} 
        \begin{tabular}{c}
          \rotatebox[origin=c]{0}{\huge P45} \\
          \rotatebox[origin=c]{0}{\huge GT}
        \end{tabular} & \hspace{-2em}
        \begin{tabular}{l}
            \subfloat{\includegraphics[width=1.3\linewidth]{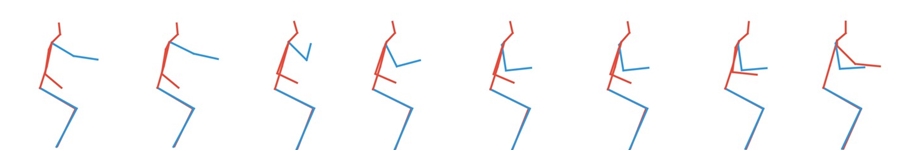}} 
        \end{tabular} \\

        \vspace{0.5em}
        \begin{tabular}{c}
          \rotatebox[origin=c]{0}{\huge P50} \\
          \rotatebox[origin=c]{0}{\huge Ours}
        \end{tabular} & \hspace{-2em}
        \begin{tabular}{l}
            \subfloat{\includegraphics[width=1.3\linewidth]{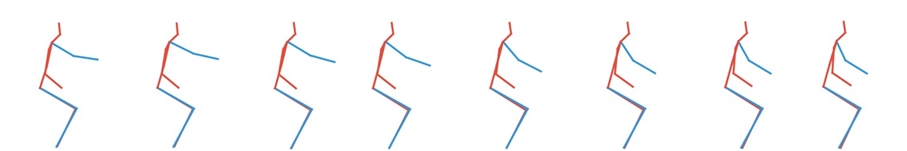}} 
        \end{tabular} & \hspace{2.5em}
        \begin{tabular}{c}
          \rotatebox[origin=c]{0}{\huge P45} \\
          \rotatebox[origin=c]{0}{\huge Ours}
        \end{tabular} & \hspace{-2em}
        \begin{tabular}{l}
            \subfloat{\includegraphics[width=1.3\linewidth]{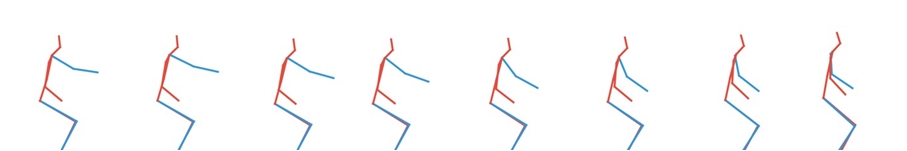}} 
        \end{tabular} \\

        \vspace{-0.2em}
        \begin{tabular}{c}
          \rotatebox[origin=c]{0}{\huge P40} \\
          \rotatebox[origin=c]{0}{\huge GT}
        \end{tabular} & \hspace{-2em}
        \begin{tabular}{l}
            \subfloat{\includegraphics[width=1.3\linewidth]{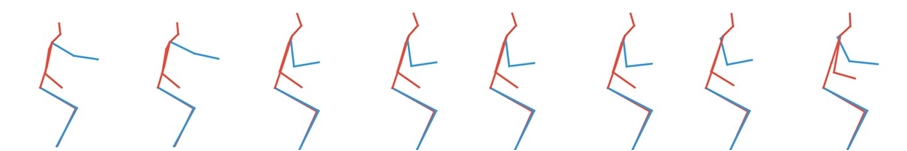}} 
        \end{tabular} & \hspace{2.5em}
        \begin{tabular}{c}
          \rotatebox[origin=c]{0}{\huge P25} \\
          \rotatebox[origin=c]{0}{\huge GT}
        \end{tabular} & \hspace{-2em}
        \begin{tabular}{l}
            \subfloat{\includegraphics[width=1.3\linewidth]{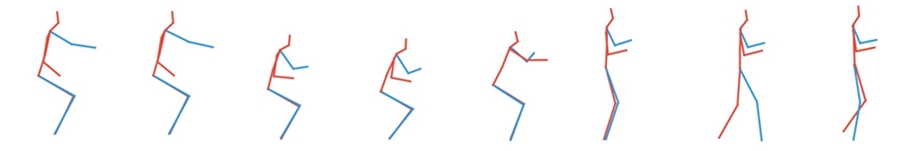}} 
        \end{tabular} \\

        \begin{tabular}{c}
          \rotatebox[origin=c]{0}{\huge P40} \\
          \rotatebox[origin=c]{0}{\huge Ours}
        \end{tabular} & \hspace{-2em}
        \begin{tabular}{l}
            \subfloat{\includegraphics[width=1.3\linewidth]{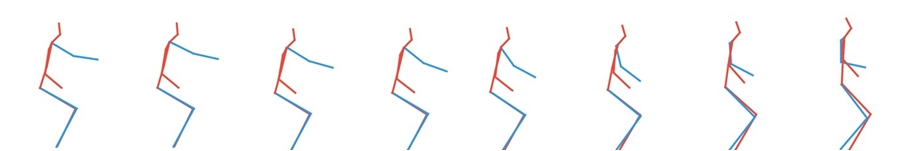}} 
        \end{tabular} & \hspace{2.5em}
        \begin{tabular}{c} \LARGE
          \rotatebox[origin=c]{0}{\huge P25} \\
          \rotatebox[origin=c]{0}{\huge Ours}
        \end{tabular} & \hspace{-2em}
        \begin{tabular}{l}
            \subfloat{\includegraphics[width=1.3\linewidth]{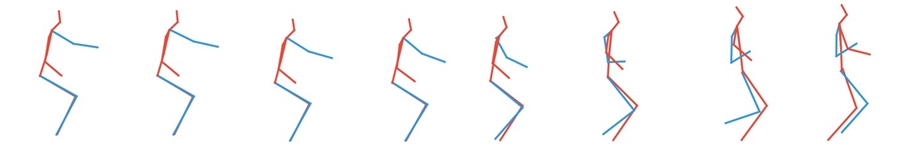}} 
        \end{tabular} \\
        
    \end{tabular}
  }
\caption{\small The visualizations for 4 different quantiles (the 50th, 45th, 40th, and 25th percentiles) from Human3.6M are presented. In each group, the poses on the top represent the ground truth and the poses on the bottom display the prediction. The threshold is set to 0.5.}
\label{fig:qualitative_quantiles}
\vspace{-0.2cm}
\end{figure*}

\subsection{Evaluation of Uncertainty Quantification} \label{SEC:EUQ}
To validate the alignment of predicted quantiles with actual quantiles, we utilize an empirical quantile evaluation metric, specifically employing ADE and FDE, which follows the diverse setup, to measure the distance between the predicted and actual quantiles. Since the true distribution is not known, we identify test samples with similar past motions using a distance threshold, and treat their subsequent movements as a proxy for the true distribution. Specifically, we order the subsequent movements in ascending order by distance and use this ordered set to determine quantiles. Given that the predicted distribution is symmetrical with the median as its most-likely motion, while the empirical distribution is skewed with the ground truth on the margin, we mirror the empirical distribution to match the form of the predicted distribution. The process of grouping empirical quantiles is consistent with the procedure used in multi-modal metrics MMADE and MMFDE. To ensure sample sizes, the sequence with pseudo futures less than 50 are excluded.

\begin{figure}[htb]\captionsetup[subfloat]{font=small}
    \centering
    \includegraphics[width=0.90\linewidth]
        {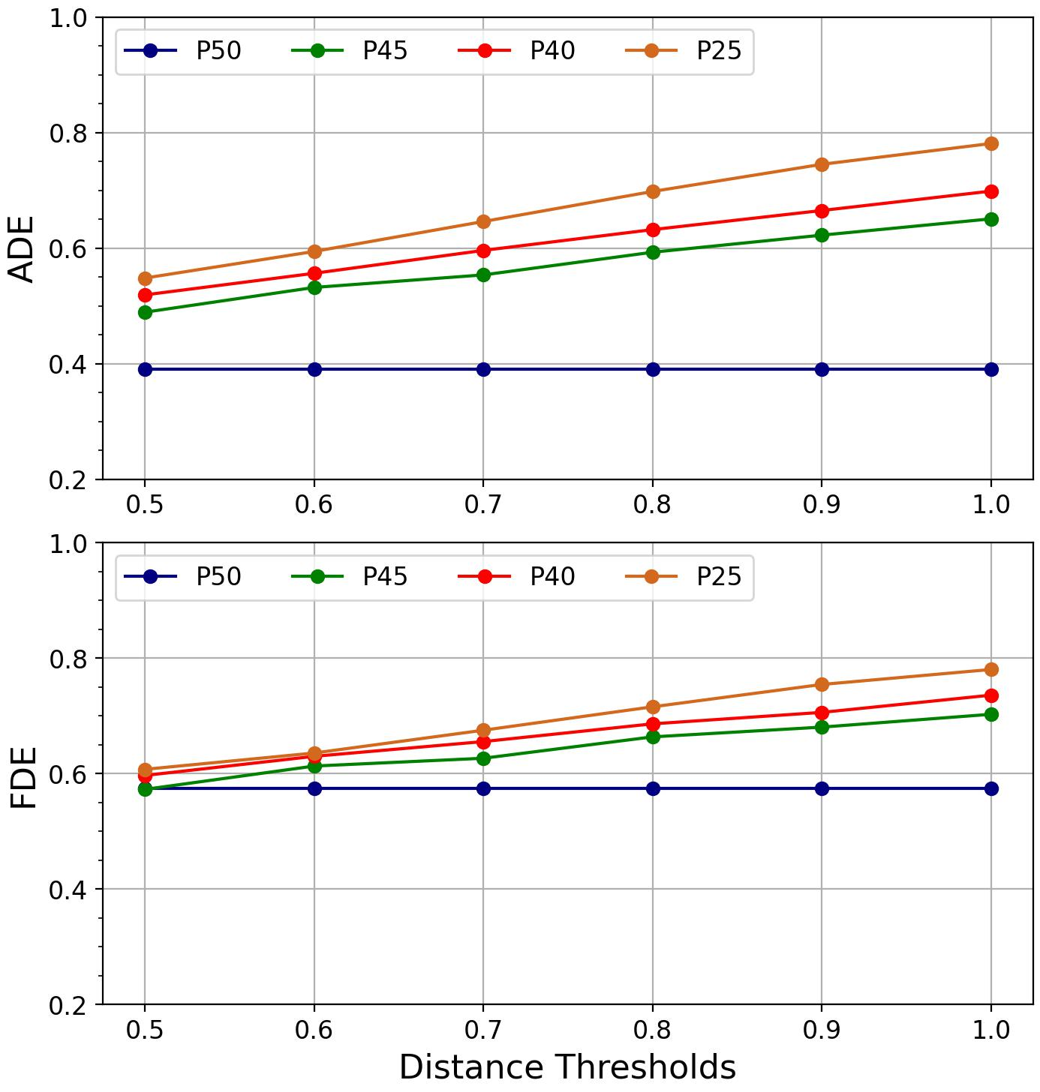}
    \caption{\small Uncertainty alignment evaluation to measure distances between the predicted quantile and the empirical quantile using ADE (top) and FDE (bottom).}
    \vspace{-0.5cm}
    \label{fig:percentile_evaluation}
\end{figure}

The quantitative results are shown in \cref{fig:percentile_evaluation}, where 4 percentiles—-50th, 45th, 40th, and 25th—-are evaluated. The 50th percentiles corresponds to the ground truth $\mathbf{X}_{T+1:T+K}$.  Low ADE and FDE, which indicate closer pose alignment and coherent trajectory over sequences important for natural appearance, are observed among all percentiles. This confirms ProbHMI accurately captures high probability regions, further supported by the qualitative results in \cref{fig:qualitative_quantiles}, where ProbHMI exhibited high fidelity in capturing movements for all percentiles. The discrepancies between predictions and empirical ground truth can be understood by two factors: 1) accumulated exposure bias and errors in long sequences, and 2) approximations in the empirical ground truth due to limited data. Despite this, our predictions still reflect movement trends, as shown in \cref{fig:qualitative_quantiles}, even though the empirical ground truth may significantly deviate from the true subsequent motion.

\subsection{Evaluation of Efficient Sampling} \label{SEC:EES}
We compare ProbHMI with baselines using much fewer samples, just 5 in our experiment, in the diverse setup to validate the sampling efficiency of ours. Here, ProbHMI employs Poisson-Disk Sampling to generate the diverse set, while other methods use a vanilla sampling schedule. The quantitative results are illustrated in \cref{TAB:sampling_efficiency}, where the value in the bracket represents the rate of change between metric values using 50 samples (shown in \cref{TAB:diverse_quan_result}) and those using 5 samples. The results show that our method not only outperforms others but also experiences only a slight performance drop (e.g., a 6.04\% increase in ADE$\downarrow$), compared to the corresponding value in \cref{TAB:diverse_quan_result}. Even in comparison with other methods with 50 samples, it is still a competitive performance, demonstrating the effectiveness of ProbHMI in estimating the future distribution with a small number of samples. In contrast, the performance of other methods degrades significantly when evaluated on 5 samples, (e.g., a 50.82\%, 49.61\% and 24.12\% increase in ADE$\downarrow$ for DLow, GSPS and HumanMAC, respectively), and all of which are much worse than any results in \cref{TAB:diverse_quan_result}.  

\begin{table}[htb]
    \renewcommand{\arraystretch}{1.5}
    \centering
    \caption{\small The evaluation using 5 samples on Human3.6M.}
    \resizebox{1.0\linewidth}{!}{
        \begin{tabular}{l|cccc}
        \toprule
            & ProbHMI & DLow & GSPS & HumanMAC \\
        \hline
            APD$\uparrow$ & 7.631(14.20\%$\uparrow$) & \textbf{16.703}(\textbf{42.26\%}$\uparrow$) & 14.801(0.29\%$\uparrow$) & 6.227(1.17\%$\downarrow$) \\
            ADE$\downarrow$ & \textbf{0.386}(\textbf{6.04\%}$\uparrow$) & 0.641(50.82\%$\uparrow$) & 0.582(49.61\%$\uparrow$) & 0.458(24.12\%$\uparrow$) \\
            FDE$\downarrow$ & \textbf{0.560}(\textbf{13.59\%}$\uparrow$) & 0.880(69.88\%$\uparrow$) & 0.783(57.86\%$\uparrow$) & 0.667(38.96\%$\uparrow$) \\
            MMADE$\downarrow$ & \textbf{0.534}(\textbf{4.50\%}$\uparrow$) & 0.701(41.61\%$\uparrow$) & 0.661(38.86\%$\uparrow$) & 0.610(19.84\%$\uparrow$) \\
            MMFDE$\downarrow$ & \textbf{0.622}(\textbf{11.46\%}$\uparrow$) & 0.889(67.42\%$\uparrow$) & 0.806(53.52\%$\uparrow$) & 0.734(34.68\%$\uparrow$) \\
        \bottomrule
        \end{tabular}
        \label{TAB:sampling_efficiency}
    }
\end{table}

\subsection{Ablation Study} \label{EQ:Ablation}
We conduct ablation studies to explore the benefit of part-aware paradigms, as shown in \cref{TAB:ablation_results}. In this context, ProbHMI w/o PAP refers to the ProbHMI model without part-aware prediction, while ProbHMI w NICE refers to ProbHMI using the standard invertible network NICE \cite{dinh2014nice}. The full version outperforms both variations across all metrics while using significantly fewer parameters, demonstrating the effectiveness of the introduced part-aware paradigm.

\begin{table}[htb]
    \renewcommand{\arraystretch}{1.25}
    \centering
    \caption{\small Experimental results of ablation studies within the diverse setup on Human3.6M.}
    \resizebox{0.8\linewidth}{!}{
        \begin{tabular}{l|cccc}
        \toprule
              & APD$\uparrow$ & ADE$\downarrow$ & FDE$\downarrow$ & FID $\downarrow$ \\
        \hline
            ProbHMI & \textbf{6.682} & \textbf{0.364} & \textbf{0.493} & \textbf{0.646} \\
            ProbHMI w/o PAP & 6.016 & 0.368 & 0.507 & 0.758 \\
            ProbHMI w/ NICE & 4.596 & 0.372 & 0.526 & 0.835 \\
        \bottomrule
        \end{tabular}
        \label{TAB:ablation_results}
    }
\end{table}

\section{Conclusion}
We present ProbHMI, a novel probabilistic framework for 3D human motion forecasting. ProbHMI addresses the limitation of prior works, specifically in quantifying uncertainty and sampling efficiency. Extensive experiments demonstrate the superiority of ProbHMI, as well as effective uncertainty quantification and calibration. To build upon ProbHMI's capabilities, incorporating stronger motion priors into the invertible network may holds promise for generating natural movements by constraining unrealistic outputs.

\newpage

\bibliographystyle{IEEEtran}
\bibliography{IEEEabrv,ref}

\end{document}